\documentclass{article}
\usepackage{kr}

\usepackage{times}
\usepackage{soul}
\usepackage{url}
\usepackage[hidelinks]{hyperref}
\usepackage[utf8]{inputenc}
\usepackage[small]{caption}
\usepackage{graphicx}
\usepackage{amsmath}
\usepackage{amsthm}
\usepackage{booktabs}
\usepackage{algorithm}
\usepackage{natbib}
\usepackage{amssymb}
\usepackage{bussproofs}
\usepackage{algpseudocode}
\usepackage{tikz}
\usetikzlibrary{shapes.geometric}
\usepackage{multirow}
\usepackage[table]{xcolor}
\usepackage{enumitem}

\graphicspath{{./img/}}

\definecolor{Acolor}{RGB}{0, 0, 0} 
\definecolor{Ecolor}{RGB}{207, 53, 53} 
\definecolor{Icolor}{RGB}{51, 51, 172} 
\definecolor{Ocolor}{RGB}{50, 138, 30} 

\definecolor{dark green}{RGB}{0,100,0} 
\definecolor{dark blue}{RGB}{3,37,126} 

\newcommand{\Achain}[2]{A#1\! - \!#2} 
\newcommand{\KB}{\mathcal{KB}} 
\newcommand{\NP}{\mathcal{P}} 
\newcommand{\Q}{\mathcal{Q}} 
\newcommand{\X}{\mathcal{X}} 
\newcommand{\T}{\tau} 
\newcommand{\D}{\triangledown} 
\newcommand{\PS}{\mathrm{M}_{\textsc{PS}}} 
\newcommand{\PBC}{\mathrm{M}_\textsc{PBC}} 
\newcommand{\Cont}{\mathcal{C}} 
\newcommand{\Fpbc}{\mathcal{S}} 

\newcommand{\msd}[2]{#1 {\footnotesize $\pm$ #2}}

\theoremstyle{definition}
\newtheorem{definition}{Definition}

\newcommand{\BI}[1]{\textbf{#1}} 
\newcommand{\DG}[1]{{\color {dark green} {#1}}}
\newcommand{\DB}[1]{{\color {dark blue} {#1}}}

\algrenewcommand{\algorithmicrequire}{\textbf{Input:}}
\algrenewcommand{\algorithmicensure}{\textbf{Output:}}

\title{Hybrid Models for Natural Language Reasoning: \\ The Case of Syllogistic Logic}

\author{%
Manuel Vargas Guzmán$^{1,3}$\and
Jakub Szymanik$^2$\and
Maciej Malicki$^1$ \\
\affiliations
$^1$University of Warsaw, Warsaw, Poland\\
$^2$University of Trento, Trento, Italy\\
$^3$Institute of Fundamental Technological Research, Polish Academy of Sciences, Warsaw, Poland\\
\emails
m.vargas-guzman@uw.edu.pl,
jakub.szymanik@gmail.com,
mmalicki@mimuw.edu.pl
}

\begin{document}

\maketitle

\begin{abstract}
  Despite the remarkable progress in neural models, their ability to generalize---a cornerstone for applications like logical reasoning---remains a critical challenge. We delineate two fundamental aspects of this ability: compositionality, the capacity to abstract atomic logical rules underlying complex inferences, and recursiveness, the aptitude to build intricate representations through iterative application of inference rules. In the literature, these two aspects are often confounded together under the umbrella term of generalization. To sharpen this distinction, we investigated the logical generalization capabilities of pre-trained large language models (LLMs) using the syllogistic fragment as a benchmark for natural language reasoning.  
  We extend classical Aristotelian syllogistic forms to build more complex structures, providing a foundational yet expressive subset of formal logic that supports controlled evaluation of essential reasoning abilities. Our findings reflect this non-trivial benchmark: while LLMs demonstrate reasonable proficiency in recursiveness, they struggle with compositionality. This disparity, however, is not uniform, as a more detailed analysis reveals variability in generalization performance across individual syllogistic types, ranging from near-perfect to significantly lower accuracy.
  To overcome these limitations and establish a reliable logical prover, we propose a hybrid architecture integrating symbolic reasoning with neural computation. This synergistic interaction enables robust and efficient inference---neural components accelerate processing, while symbolic reasoning ensures completeness. Our experiments show that high efficiency is preserved even with relatively small neural components. As part of our proposed methodology, this analysis gives a rationale and highlights the potential of hybrid models to effectively address key generalization barriers in neural reasoning systems.
\end{abstract}

\section{Introduction}

Neural models have achieved substantial advancements at an accelerated pace in recent years. 
However, they continue to face challenges in generalizing---a capability that is crucial for tasks such as logical deduction.
While they excel at pattern recognition, these models often struggle with the systematicity and robustness required for sound reasoning 
\citep{marcus2018deeplearningcriticalappraisal, Lake_Ullman_Tenenbaum_Gershman_2017, huang-chang-2023-towards, DBLP:journals/corr/abs-2404-01869}.
This is particularly evident in their limited capacity to generalize beyond the training data, especially in tasks that demand a deep understanding of compositional structures \citep{hupkes2023taxonomy}.

In this work, we focus on two fundamental and complementary aspects of generalization in the context of logical reasoning: compositionality and recursiveness. \emph{Compositionality} is the ability to recognize that the logical rules governing simple inferences remain valid when embedded within more complex structures. While the Principle of Compositionality \citep{janssen1997compositionality} is traditionally framed as a ``bottom-up'' process—where the meaning of a complex expression is determined by the meanings of its constituent parts, we diverge from this standard usage to emphasize an analytical, ``top-down'' perspective. Under our definition, a system is compositional if it can abstract atomic logical rules from complex structures and apply them to its simpler components. For example, if a model can derive the inference ``if all $a$ are $b$, all $b$ are $c$, and all $c$ are $d$, then all $a$ are $d$,'' it should also be able to derive the constituent inference ``if all $a$ are $b$ and all $b$ are $c$, then all $a$ are $c$.'' \emph{Recursiveness}, on the other hand, is the ability to construct complex representations through the iterative application of a finite set of rules. While recursiveness is often viewed as the constructive mechanism of compositionality, we treat it as a distinct property to isolate a model’s capacity for length-expansion. For instance, a recursive model can extend the transitive inference ``if all $a$ are $b$ and all $b$ are $c$, then all $a$ are $c$'' to a longer chain: ``if all $a$ are $b$, all $b$ are $c$, and all $c$ are $d$, then all $a$ are $d$.'' A system that is merely recursive might produce longer chains by mimicking surface-level patterns, but a truly compositional system demonstrates an understanding of the argument's structure by successfully reasoning about its sub-parts. It is therefore possible for a model to be recursive—processing arbitrarily long chains—without being compositional, meaning it lacks an understanding of how individual links in the chain combine. This lack of analytical compositionality remains a key limitation of current neural models \citep{guzman-etal-2024-testing, Lake:2023aa, bertolazzi-etal-2026-teaching}.

To investigate this issue, we examine the logical generalization capabilities of large pre-trained language models (LLMs) using syllogistic logic—a well-defined, yet non-trivial, fragment of natural language that captures a fundamental aspect of human reasoning. 
Syllogistic logic was chosen as a clearly tractable baseline for compositional and recursive generalization, as logics that are too expressive---such as full first-order logic---are computationally intractable.
We fine-tuned LLMs on two distinct reasoning tasks to support the construction of direct and indirect formal proofs. 
Specifically, for a given knowledge base $\KB$ and a hypothesis $H$, the first task is \emph{premise selection}, i.e., predicting a minimal subset $\NP \subseteq \KB$ such that $\NP \vdash H$; 
the second task is \emph{proof by contradiction}, i.e., predicting a formula $F$ such that $\KB \cup \{ \overline{H} \} \vdash F$ and $\KB \vdash \overline{F}$.
These tasks were designed to probe different facets of logical generalization, with premise selection requiring an understanding of the relationships between statements and proof by contradiction testing the ability to reason about counterfactuals and derive logical consequences. 
To ensure a rigorous evaluation, we trained and tested on multiple knowledge bases generated from controlled synthetic data, which incorporates pseudowords to avoid content bias \citep{bertolazzi-etal-2024-systematic}.    
Our experiments reveal a significant disparity: while LLMs demonstrate reasonable proficiency in recursive reasoning, they struggle with compositional generalization. Specifically, when trained on simpler inferences, LLMs can recognize analogous simple inferences across different knowledge bases and generalize to a certain extent to more complex inference patterns. However, models trained exclusively on complex inferences exhibit a substantial performance drop when required to identify the underlying simpler components.
Importantly, we observe systematic variability in performance across different reasoning structures, indicating that generalization is type-dependent. Models generalize certain forms consistently, yet struggle with others.
That these phenomena arise already within syllogistic logic is theoretically significant. 
Despite its simplicity, this fragment reveals a clear structural boundary between inference types that LLMs can generalize and those they cannot. 

\begin{figure} 
	\centering
	\tikzset{>=latex}
	\scalebox{0.83}{\begin{tikzpicture}
			\node[cylinder, 
			draw = dark blue, 
			cylinder uses custom fill, 
			cylinder body fill = dark blue!10, 
			cylinder end fill = dark blue!40,
			aspect = 0.3, 
            line width = 1pt,
			minimum height = 1.2cm,
			minimum width = 1.5cm,
			shape border rotate = 90] (KB) at (0,2) {$\KB$};		
			\node[rectangle,
			draw = dark blue,
			fill = dark blue!10,
            line width = 1pt,
			minimum width = 1.2cm] (H) at (0,0.5) {$H$};		
			\node[rectangle,
			draw=dark green,
            line width = 1pt,
			minimum height = 3.8cm,
			minimum width = 5.1cm, 
            dashed] (HM) at (3.9, 0.8) {};
			\node[rectangle,
			draw = dark green,
            line width = 1.2pt,
			minimum height = 1.4cm,
			minimum width = 4.65cm, 
            fill = dark green!10,
            rounded corners = 4pt] (CC) at (3.9,-0.2) {};		
			\node[rectangle,
			draw=dark green,
			minimum height=0.7cm,
			minimum width=1.7cm,
			fill = dark green!10,
            rounded corners = 4pt] (PS) at (2.6,0) {}; 
			\node[rectangle,
			draw=dark green,
			minimum height=0.7cm,
			minimum width=2.4cm, 
			fill = dark green!10,
            rounded corners = 4pt] (PBC) at (4.85, 0) {};
			\node[rectangle,
			draw = dark green,
			fill = dark green!10,
            line width = 1.2pt,
			minimum height = 1cm,
			minimum width = 3.6cm,
            rounded corners = 4pt] (SC) at (3.9,2) {\BI{\textsc{symbolic prover}}};		
			\node[rectangle,
			draw = dark blue,
			fill = dark blue!10,
            line width = 1pt] (O) at (8,2) {$\D: \KB \vdash H$}; 
			\node at (2.85, 1) {$\NP$}; 
			\node at (5.1, 1) {$F$};
			\node at (4, -0.6) {\BI{\textsc{neural assistant}}};
			\node at (2.6, 0.15) {\small \textsc{premise}}; 
			\node at (2.6, -0.15) {\small \textsc{selection}}; 
			\node at (4.85, 0.15) {\small \textsc{proof by}};
			\node at (4.85, -0.15) {\small \textsc{contradiction}};
			\draw[->, line width=1.2pt] (KB.east) -- (SC.west);
			\draw[line width=1.2pt] (1.2,0.5) -- (1.2,2);		
			\draw[line width=1.2pt] (H.east) -- (1.21,0.5);
			\draw[line width=1.2pt] (1.8,1.75) -- (2.1,1.75);
			\draw[->, line width=1.2pt] (1.8,1.75) -- (1.8,0.45);
			\draw[->, line width=1.2pt] (PS.north) -- (2.6, 1.5); 
			\draw[->, line width=1.2pt] (PBC.north) -- (4.85,1.5);
			\draw[->, line width=1.2pt] (SC.east) -- (O.west);
		\end{tikzpicture}       
	}
	\caption{Overview of the hybrid architecture. \DB{Input:} a knowledge base $\KB$ and a hypothesis $H$. \DG{Hybrid model:} the neural models assist the symbolic prover by providing a subset $\NP \subseteq \KB$ such that $\NP \vdash H$, and a refutation formula $F$ such that $\KB \cup \{ \overline{H} \} \vdash F$ and $\KB \vdash \overline{F}$. \DB{Output:} a proof $\D$ of $H$ from $\KB$.}
	\label{fig:hybrid_architecture}        
\end{figure}        

To address this limitation, we propose a new research program consisting of two elements. On the theoretical side, we aim to understanding how different reasoning building blocks interact with deep-learning model performance on generalization tasks. On the practical side, we develop a hybrid architecture 
(see Figure \ref{fig:hybrid_architecture}) that integrates the pattern-matching strengths of neural networks with the formal rigor and completeness of symbolic reasoning. In this framework, the neural component serves as an auxiliary to the symbolic prover, efficiently providing candidate premises and formulas to guide the derivation process.
Furthermore, to evaluate the impact of the assistant on the symbolic prover in terms of the number of computational steps required during proof search, we implemented a baseline non-deterministic prover.
This synergistic approach aims to overcome the limitations of purely neural approaches by enforcing logical consistency and enabling systematic generalization.

The key contributions of this work are: (1) A rigorous empirical demonstration that, despite their recursive capabilities, LLMs lack true compositionality, a crucial requirement for robust logical reasoning. We emphasize the importance of distinguishing between these two properties in evaluations of neural reasoning systems. 
(2) A theoretical finding that generalization in neural models is constrained by logical structure. This highlights that certain inference types can be effectively learned while others pose difficulties, providing a systematic analysis for understanding the limits of neural reasoning.
(3) A hybrid approach that leverages neural networks for efficient inference while relying on symbolic reasoning to guarantee logical completeness and correctness. 
We evaluate the robustness of the hybrid model, focusing on how neural assistance affects symbolic proof search across different model sizes and configurations.

We present our hybrid framework in three parts. The first part establishes the symbolic foundation of the approach: we introduce the underlying syllogistic logic (Section~\ref{sec:proof_system}) and a non-deterministic algorithm that implements a prover (Section~\ref{sec:symbolic_component}). The second part focuses on the connectionist setting, where we train neural models to approximate this reasoning process. We describe the synthetic dataset and fine-tuning setup (Section~\ref{sec:connectionist_component}), and then analyze how these models generalize beyond their training distribution, relating performance to structural properties of syllogistic proofs (Section~\ref{sec:analysis_of_generalization}). The third part brings both perspectives together by examining how the symbolic and neural components interact within a unified experimental setting (Section~\ref{sec:hybrid_models}).

\section{A Syllogistic Proof System}
\label{sec:proof_system}	
We adopt a syllogistic proof system based on \citet{Smiley1973-SMIWIA} as a foundation for implementing hybrid reasoning models. 
The system comprises four formula types: $A a b$, $E a b$, $I a b$, and $O a b$, corresponding to the natural-language forms ``All $a$ are $b$'', 
``No $a$ are $b$'', ``Some $a$ are $b$'', and ``Some $a$ are not $b$.'' 
This framework extends the classical two-premise syllogism by allowing derivations from arbitrarily many premises. 
In particular, $A$-formulas are treated as chains representing transitive inclusion among terms. 
These chains provide the structural backbone of syllogistic reasoning, enabling its compositional and recursive analysis.

\subsection{Syntax and Semantics}
We formally define the syntax and semantics of a language with four \emph{quantifier} symbols $\Q = \{ A, E, I, O \}$ 
and an infinite set of \emph{term} symbols $\X = \{ a, b, c, \ldots \}$ denoted by lowercase letters (sometimes with subscripts).
\emph{Well-formed formulas} over this language are built as $A a b$, $E a b$, $I a b$, or $O a b$.	
An \emph{A-chain}, denoted as $\Achain{a}{b}$, represents either the formula $A a b$ or the sequence of two or more formulas 
$A a c_1, A c_1 c_2, \ldots, Ac_{n-1} c_n, Ac_n b$ (for $n \geq 1$). 	
In what follows, when we refer to a formula we mean a well-formed formula.
Moreover, we use capital letters (e.g., $F$ or $H$) to denote formulas, and capital calligraphic letters 
(e.g., $\KB$, $\mathcal{F}$ or $\mathcal{P}$) to denote sets of formulas, unless stated differently.

The meaning of syllogistic formulas can be defined using set-theoretic relationships. 
Let us interpret terms $a$ and $b$ as non-empty subsets of an underlying universe $M$; then
$A a b$ is true \emph{iff} $a \subseteq b$; $E a b$ is true \emph{iff} $a \cap b = \emptyset$; $I a b$ is true \emph{iff} $a \cap b \neq \emptyset$;
and $O a b$ is true \emph{iff} $a \not\subseteq b$.
A set of syllogistic formulas $\mathcal{F}$ is \emph{consistent} if there exists an interpretation (i.e., a universe $M$ and an assignment of subsets of $M$ to terms) under which every $F \in \mathcal{F}$ is true; otherwise it is \emph{contradictory}. 

Note that $A a b$ and $O a b$ are contradictory, and the same about $I a b$ and $E a b$.
We denote the negation of a formula $F$ as $\overline{F}$, i.e.,
$\overline{A a b} = O a b$, $\overline{O a b} = A a b$, $\overline{I a b} = E a b$, and $\overline{E a b} = I a b$.
Last but not least, $I$- and $E$-formulas are symmetrical. That is, $I a b$ and  $I b a$ have the same meaning, and so do $E a b$ and $E b a$.

\subsection{Inference and Proofs}
We define a syllogism as an inference from a set of premises, referred to as a knowledge base, to a conclusion. 

\begin{definition}[Proof]
	\label{def:proofs}
	The following is a mutually recursive definition to characterize formal proofs from a set of formulas using tree notation.
	A \emph{proof} $\D$ of a hypothesis $H$ from a set of premises $\mathcal{F}$ is one of the following three types:
	\begin{enumerate}[label=(\roman*), leftmargin=*, itemsep=0.5em]
		\item \emph{Trivial proof}: for every $H \in \mathcal{F}$, the following tree is a proof of $H$ from $\mathcal{F}$
		\begin{prooftree}
			\AxiomC{}
			\RightLabel{\scriptsize(i)}
			\UnaryInfC{$H$}
		\end{prooftree}
        
		\item \emph{Rule-based proofs}: for $\D'$ and $\D''$ that are proofs from $\mathcal{F}$, the following trees are proofs from $\mathcal{F}$  \\	
            \noindent
            \begin{minipage}{0.48\linewidth}
            \centering
            \begin{prooftree}
              \AxiomC{}
              \noLine
              \UnaryInfC{$\D'$}
              \noLine
              \UnaryInfC{$A a b$}
              \AxiomC{}
              \noLine
              \UnaryInfC{$\D''$}
              \noLine
              \UnaryInfC{$A b c$}
              \RightLabel{\scriptsize(r1)}
              \BinaryInfC{$A a c$}
            \end{prooftree}
            \end{minipage}
            \hfill
            \begin{minipage}{0.48\linewidth}
            \centering
            \begin{prooftree}
              \AxiomC{}
              \noLine
              \UnaryInfC{$\D'$}
              \noLine
              \UnaryInfC{$A a b$}
              \AxiomC{}
              \noLine
              \UnaryInfC{$\D''$}
              \noLine
              \UnaryInfC{$E b c$}
              \RightLabel{\scriptsize(r2)}
              \BinaryInfC{$E a c$}
            \end{prooftree}
            \end{minipage}
            
            \vspace{2ex}
            
            \noindent
            \begin{minipage}{0.48\linewidth}
            \centering
            \begin{prooftree}
              \AxiomC{}
              \noLine
              \UnaryInfC{$\D'$}
              \noLine
              \UnaryInfC{$E b a$}
              \RightLabel{\scriptsize(r3)}
              \UnaryInfC{$E a b$}
            \end{prooftree}
            \end{minipage}
            \hfill
            \begin{minipage}{0.48\linewidth}
            \centering
            \begin{prooftree}
              \AxiomC{}
              \noLine
              \UnaryInfC{$\D'$}
              \noLine
              \UnaryInfC{$A b a$}
              \RightLabel{\scriptsize(r4)}
              \UnaryInfC{$I a b$}
            \end{prooftree}
            \end{minipage}
			
		\item \emph{Proof by contradiction}: for $\D'$ that is a proof from $\mathcal{F} \cup \nobreak\{ \overline{H} \}$ and $\D''$ that is a proof from $\mathcal{F}$, the following tree is a proof from $\mathcal{F}$ 
			\begin{prooftree}
				\AxiomC{}
				\noLine
				\UnaryInfC{$\D'$}
				\noLine
				\UnaryInfC{$F$}
				\AxiomC{}
				\noLine
				\UnaryInfC{$\D''$}
				\noLine
				\UnaryInfC{$\overline{F}$}
				\RightLabel{\scriptsize(iii)}
				\BinaryInfC{$H$}
			\end{prooftree}
	\end{enumerate}
\end{definition}	

\begin{definition}[Inference]
	Let $\mathcal{F}$ be a set of formulas (\emph{premises}) and let $F$ be a formula (\emph{conclusion}).
	We write $\mathcal{F} \vdash F$, to denote that $F$ is provable (or derivable) from $\mathcal{F}$, i.e., there exists a proof of $F$ from $\mathcal{F}$.
\end{definition}

\paragraph{Soundness and Completeness} 
The deductive system presented above is based on the framework developed by \cite{Smiley1973-SMIWIA}. 
In this system, a formula is provable if and only if it is true in all interpretations (as established in Theorems 3 and in 4). 
It therefore follows that the system is both sound and complete.

\begin{definition}[Minimal Inference] 
    An inference $\mathcal{F} \vdash F$ is \emph{minimal} if for no proper subset $\mathcal{F}' \subsetneq \mathcal{F}$, it is the case that $\mathcal{F}' \vdash F$.
\end{definition}

Minimal inferences are essential to design a connectionist prover assistant, as they use only the necessary premises to derive a conclusion.
Table \ref{tab:syllogisms} depicts all types of minimal syllogistic inferences, denoted by $\T$, as outlined in \cite{guzman-etal-2024-testing}.	

\begin{table} 
	\centering	
	\scalebox{0.9}{\begin{tabular}{ c l }
			\toprule
			$\T$ & \textbf{Syllogism} \\
			\midrule
			1 & $\{ \Achain{a}{b}, \Achain{c}{d}, O a d \} \vdash O b c$ \\
			2 & $\{ \Achain{a}{b} \} \vdash A a b$ \\
			3 & $\{ \Achain{a}{b}, \Achain{c}{d}, \Achain{a}{e}, E d e \} \vdash O b c$ \\
			4 & $\{ \Achain{a}{b}, \Achain{a}{c} \} \vdash I b c$ \\
			5 & $ \{ \Achain{a}{b}, \Achain{c}{d}, \Achain{e}{f}, I a e, E d f \} \vdash O b c$  \\
			6 & $ \{ \Achain{a}{b}, \Achain{c}{d}, E b d \} \vdash E a c $ \\
			7 & $ \{ \Achain{a}{b}, \Achain{c}{d}, I a c \} \vdash I b d $ \\
			\bottomrule
	\end{tabular}}
    \caption{Types of syllogistic inferences}
	\label{tab:syllogisms}
\end{table}	

\section{Symbolic Component}
\label{sec:symbolic_component}
We implemented a basic automated syllogistic prover that takes as input a knowledge base $\KB$ and a hypothesis $H$, 
i.e., a proposed conclusion to be derived from $\KB$. The overall procedure is described in Algorithm \ref{alg1:syllogistic_prover}.
The prover first attempts to construct a proof using types (i) and (ii) via the \textsc{drv} function. 
If this attempt fails, it proceeds to use proof by contradiction (type (iii)) through the \textsc{pbc} function.
Successful partial derivations are stored in a set $\Delta$ as triples that correspond to proof steps: a set of premises 
$\mathcal{F}$, a derived formula $F$, and the proof type by which the derivation is obtained.
If $H$ is valid, the prover returns its corresponding derivation steps (line 3). 
Otherwise, it exhaustively explores all possible derivations before concluding that $H$ is invalid.

\begin{algorithm}
	\caption{Syllogistic Prover}
	\label{alg1:syllogistic_prover}
	\begin{algorithmic}[1]
		\Require A hypothesis $H$ and a knowledge base $\KB$
		\Ensure A proof $\D$ of $H$ from $\KB$ (if $H$ is valid)
		\State $\Delta \gets \emptyset$ 
		\If{\Call{drv}{$H, \KB$} \textbf{or} \Call{pbc}{$H, \KB$}}
		\State $\D \gets$ \Call{get\_steps}{$H, \Delta$} 
		\State\Return{$\D$}
		\Else
		\State\Return{\texttt{False}}
		\EndIf
	\end{algorithmic}
\end{algorithm}

\subsection{Derivation of Formulas}
The central component of the prover is the recursive function \textsc{drv}, described in Algorithm \ref{alg2:derive}. 
The function takes as input the knowledge base $\KB$ and the hypothesis $H$, and initially checks whether proof of type (i) can be directly applied. 
If the base case is not satisfied, the function attempts to derive $H$ by non-deterministically searching for a set of formulas $\mathcal{F}$
such that $H$ follows from $\mathcal{F}$ (line 4) by using rules of inference, i.e., proofs of type (ii). Then the process is applied recursively to each $F \in \mathcal{F}$, 
continuing until the base case is met or no further applicable rules and formula combinations are available. 
To prevent redundant computations and potential infinite loops, the algorithm is optimized by storing partial proofs (lines 2 and 5) whenever the base case is reached.
These stored derivations are reused if the same inputs are encountered again. Additionally, the system tracks failed derivation attempts to avoid repeating them in subsequent searches.
Nevertheless, this search process remains computationally demanding, particularly when constructing \emph{A-chains}, 
where the algorithm conducts a non-deterministic search over formulas of the form $Aab$, generated using terms drawn from the knowledge base. 
For each step, there are up to $n(n-1)$ candidates, where $n$ denotes the number of distinct terms in the knowledge base, 
with each choice leading to further alternatives and causing the search space to grow rapidly.

\begin{algorithm}
	\caption{Derive (DRV)}
	\label{alg2:derive}
	\begin{algorithmic}[1]
        \Require A hypothesis $H$ and a knowledge base $\KB$
        \Ensure \texttt{True} iff $H$ is provable 
		\If{$H \in \KB$} 
		\State $\Delta$ $\gets$ $\Delta \cup \{ (\emptyset, H, \text{(i)}) \}$ 
		\State\Return{\texttt{True}}
		\ElsIf{\Call{is\_derivable}{$\mathcal{F}, H$}} 
		\State $\Delta$ $\gets$ $\Delta \cup \{ (\mathcal{F}, H, \text{(ii)}) \}$ 
		\ForAll{$F \in \mathcal{F}$}
		\State\Return\Call{drv}{$F, \KB$} 
		\EndFor
		\Else
		\State\Return{\texttt{False}}
		\EndIf
	\end{algorithmic}	
\end{algorithm}

\subsection{Proof by Contradiction}
The final component of the prover, denoted \textsc{pbc}, is specified in Algorithm \ref{alg3:pbc}. This function gets the same inputs as \textsc{drv}, and it aims to 
prove the hypothesis $H$ by contradiction by finding a formula $F$ such that $\KB \cup \{ \overline{H} \} \vdash F$ and $\KB \vdash \overline{F}$.
The algorithm begins by generating all possible contradictory formula pairs $(F, \overline{F})$,
where $F$ ranges over all formulas obtained by applying quantifiers from $\Q$ to pairs of terms drawn from $\X$, with $\X$ restricted to the terms occurring in $\KB$.
It then iterates through these pairs in search of a contradiction. 
If such proof is found, it is stored; otherwise, the process continues until all pairs have been exhausted.
The search for candidate pairs is performed in a non-deterministic manner, which can result in significant computational cost, 
as the algorithm calls the \textsc{drv} function for each formula within every potential pair.

\begin{algorithm}
	\caption{Proof by Contradiction (PBC))}
	\label{alg3:pbc}
	\begin{algorithmic}[1]			
        \Require A hypothesis $H$ and a knowledge base $\KB$
        \Ensure \texttt{True} iff $H$ is provable by contradiction
        \State $\Cont \gets$ \Call{all\_pairs}{$\Q, \X$} 
		\ForAll{$(F, \overline{F}) \in \Cont$} 
        \If{ \Call{drv}{$\overline{F}, \KB$} \textbf{and} \Call{drv}{$F, \KB \cup \{ \overline{H} \}$} }
		\State $\Delta$ $\gets$ $\Delta \cup \{ (\{ F, \overline{F} \}, H, \text{(iii)}) \}$
		\State\Return{\texttt{True}}			
		\EndIf
		\EndFor
		\State\Return{\texttt{False}}
	\end{algorithmic}	
\end{algorithm}

\section{Connectionist Component}
\label{sec:connectionist_component}
Our hybrid models integrate two distinct connectionist components designed to support the prover: one assists in selecting the necessary premises required to derive the hypothesis, 
while the other facilitates the identification of a suitable formula for proof by contradiction. 
To develop these models, we generated synthetic training data and fine-tuned pre-trained large language models (LLMs) on the specific tasks assigned to each component.
The code for reproducing the experiments is publicly available at: \url{https://github.com/manuel-vg/neurosylogix}.

\subsection{Synthetic Data}	
A knowledge base $\KB$ can be formally represented as an edge-labeled graph $G = (V, E, \gamma)$, where the set of vertices $V$ are terms from the domain $\X$
and the set of formulas correspond to the set of edges $E \subseteq \{ (u,v) \;|\; u,v \in V \text{ and } u \neq v \}$ along with a labeling function $\gamma: E \mapsto \Q$ 
that maps edges to syllogistic quantifiers (see Figure \ref{fig:graph_KB} for an example).
We produced synthetic knowledge bases by randomly generating graphs such that the resulting knowledge bases are consistent. 
Furthermore, we imposed the constraint that for every formula $F$ derivable from a given knowledge base $\KB$, there exists a unique subset $\NP \subseteq \KB$ such that $\NP \vdash F$ is minimal. 	
This property is critical for eliminating redundant derivations of the same hypothesis. We refer to such structures as \emph{non-redundant} knowledge bases.

\begin{figure}[!t]
	\centering
	\tikzset{>=latex}
	\begin{minipage}{0.65\linewidth}
		\scalebox{0.8}{\begin{tikzpicture}
				\node (x1) at (0,0) {$x_1$};
				\node (x2) at (0,1.5) {$x_2$};
				\node (x3) at (0,3) {$x_3$};
				\node (x4) at (-1,4.5) {$x_4$};
				\node (x5) at (1,4.5) {$x_5$};
				\node (x6) at (4.5,0) {$x_6$};
				\node (x7) at (4.5,1.5) {$x_7$};
				\node (x8) at (3.5,3) {$x_8$};
				\node (x9) at (5.5,3) {$x_9$};
				\node (x10) at (3.5,4.5) {$x_{10}$};
				\node (x11) at (5.5,4.5) {$x_{11}$};
				
				\draw[thick, Acolor, ->] (x1) -- (x2) node[midway,left] {$A$};
				\draw[thick, Acolor, ->] (x2) -- (x3) node[midway,left] {$A$};
				\draw[thick, Acolor, ->] (x3) -- (x4) node[midway,left] {$A$};
				\draw[thick, Acolor, ->] (x3) -- (x5) node[midway,left] {$A$};			
				\draw[thick, Acolor, ->] (x6) -- (x7) node[midway,left] {$A$};
				\draw[thick, Acolor, ->] (x7) -- (x8) node[midway,left] {$A$};
				\draw[thick, Acolor, ->] (x7) -- (x9) node[midway,right] {$A$};
				\draw[thick, Acolor, ->] (x8) -- (x10) node[midway,left] {$A$};
				\draw[thick, Acolor, ->] (x9) -- (x11) node[midway,right] {$A$};
				
				\draw[thick, Ecolor, <->] (x4) to[bend left] (x11);
				\node (l2) at (2.2,5.3) {\color{Ecolor} $E$};
				
				\draw[thick, Icolor, <->] (x1) -- (x8) node[midway, above] {$I$};
				
				\draw[thick, Ocolor, ->] (x5) to[bend left] (x3);
				\node (l3) at (1.1,3.6) {\color{Ocolor} $O$};						
				\draw[thick, Ocolor, ->] (x4) -- (x5) node[midway, above] {$O$};
				\draw[thick, Ocolor, ->] (x11) to[bend right] (x7);
				\node (l3) at (4.4,3.5) {\color{Ocolor} $O$};
		\end{tikzpicture}}
	\end{minipage}\hfill
	\begin{minipage}{0.30\linewidth}
		\scalebox{0.8}{\begin{tabular}{ l }
				$\KB = \{ Ax_1x_2,$ \\
                \hspace{1.08cm} $Ax_2x_3,$ \\
                \hspace{1.08cm} $Ax_3x_4,$ \\
                \hspace{1.08cm} $Ax_3x_5,$ \\
                \hspace{1.08cm} $Ax_6x_7,$ \\
                \hspace{1.08cm} $Ax_7x_8,$ \\
				\hspace{1.08cm} $Ax_7x_9,$ \\
                \hspace{1.08cm} $Ax_8x_{10},$ \\
                \hspace{1.08cm} $ Ax_9x_{11},$ \\
                \hspace{1.08cm} ${\color{Ecolor} Ex_4x_{11}},$ \\
				\hspace{1.08cm} ${\color{Icolor} Ix_1x_8},$ \\
                \hspace{1.08cm} ${\color{Ocolor} Ox_4x_5,}$ \\
                \hspace{1.08cm} ${\color{Ocolor} Ox_5x_3,}$ \\
                \hspace{1.08cm} ${\color{Ocolor} Ox_{11}x_7}\}$
		\end{tabular}}
	\end{minipage}

    \begin{center}
        \scalebox{0.9}{\begin{tabular}{ c l }
        \toprule
		$\T$ & \textbf{Examples of syllogisms} \\
		\midrule
        1 & $\{ \Achain{x_1}{x_3}, Ox_5x_3 \} \vdash Ox_5x_1$ \\[0.1cm]
        2 & $\{ \Achain{x_6}{x_{11}} \} \vdash Ax_6x_{11}$ \\[0.1cm]
        3 & $\{ \Achain{x_2}{x_4}, \Achain{x_7}{x_{11}}, Ax_7x_8, Ex_4x_{11} \} \vdash Ox_8x_2$ \\[0.1cm]
        4 & $\{ \Achain{x_7}{x_{10}}, \Achain{x_7}{x_{11}} \} \vdash Ix_{10}x_{11}$ \\[0.1cm]
        5 & $\{ \Achain{x_1}{x_4}, \Achain{x_6}{x_{11}}, Ax_8x_{10}, Ex_4x_{11}, Ix_8x_1 \} \vdash Ox_{10}x_6$ \\[0.1cm]
        6 & $\{\Achain{x_1}{x_4}, \Achain{x_6}{x_{11}}, Ex_4x_{11} \} \vdash Ex_6x_1$ \\[0.1cm]
        7 & $\{ \Achain{x_1}{x_4}, Ax_8x_{10}, Ix_8x_1 \} \vdash Ix_4x_{10}$ \\
        \bottomrule
    \end{tabular}}
    \end{center}
	\caption{Example of a knowledge base $\KB$ represented as a graph, along with valid inferences that can be derived from $\KB$.}
	\label{fig:graph_KB}
\end{figure}

To convert these structured representations into natural language inputs, terms are replaced with artificially generated pseudowords, 
and formulas are translated into a textual representation (see Figure \ref{fig:substitutions} for an example). 	

\begin{figure}[!t]
	\centering
	\tikzset{>=latex}
	\scalebox{0.8}{\begin{tikzpicture}
			\draw (-1.2,-0.6) rectangle (3.8,3.6);
			
			\node (x1) at (0,0) {$x_1$};
			\node (x2) at (0,1.5) {$x_2$};
			\node (x3) at (0,3) {$x_3$};
			\node (x4) at (1.5,0) {$x_4$};
			\node (x5) at (1.5,1.5) {$x_5$};
			\node (x6) at (3,0) {$x_6$};
			\node (x7) at (3,1.5) {$x_7$};
			\node (x8) at (3,3) {$x_8$};
			
			\draw[thick, Acolor, ->] (x1) -- (x2) node[midway,right] {$A$};
			\draw[thick, Acolor, ->] (x2) -- (x3) node[midway,right] {$A$};
			\draw[thick, Acolor, ->] (x4) -- (x5) node[midway,right] {$A$};
			\draw[thick, Acolor, ->] (x6) -- (x7) node[midway,right] {$A$};			
			\draw[thick, Acolor, ->] (x7) -- (x8) node[midway,right] {$A$};
			
			\draw[thick, Ecolor, <->] (x3) -- (x8) node[midway, above] {$E$};
			
			\draw[thick, Icolor, <->] (x1) -- (x4) node[midway, below] {$I$};
			
			\draw[thick, Ocolor, ->] (x3) to[bend right] (x1);
			\node (l1) at (-0.8,1.5) {\color{Ocolor} $O$};						
			
            \node[draw, thick, align=center, rounded corners] (ftc) at (4, -1.5) {\textsc{symbolic-text interface} };
			
            \node[draw, fill={rgb:black,1;white,7}, text width=6cm] (subs) at (4, -2.8) { \color{black}{$ [x_1 / \text{preac}, \; x_2 / \text{verde}, \; x_3 / \text{usni}, \; x_4 / \text{goed},$ \\[0.15cm] $\; x_5 / \text{itil}, \; x_6 / \text{entpi}, \; x_7 / \text{ondy}, \; x_8 / \text{ramer}] $} };
			
            \node[draw, text width=4.2cm] at (7, 1.5) {\\[0.05cm] $\{$\text{All preac are verde}, \\[0.1cm] \text{  All verde are usni}, \\[0.1cm] \text{ All goed are itil}, \\[0.1cm] \text{ All entpi are ondy}, \\[0.1cm] \text{ All ondy are ramer}, \\[0.1cm] 
				{\color{Ecolor} \text{ No usni are ramer}}, \\[0.1cm] {\color{Icolor} \text{ Some preac are goed}}, \\[0.1cm] {\color{Ocolor} \text{ Some usni are not preac}}$\}$};
			
			\draw[thick, <->] (2, -0.6) -- (2, -1.3);
			\draw[thick, <->] (6, -0.6) -- (6, -1.3);					
			\draw[thick, ->] (subs) -- (ftc) {};
	\end{tikzpicture}}		
    \caption{Example of a symbolic--text interface mapping a set of syllogistic formulas, represented as a graph, to a set of natural-language formulas via pseudoword substitution, and vice versa.}

	\label{fig:substitutions}
\end{figure}

\subsection{Fine-Tuning LLMs}
To evaluate the impact of neural assistance on symbolic proof search, we fine-tuned two transformer-based architectures of different sizes:
FLAN-T5-base \citep{JMLR:v21:20-074}, a compact \emph{encoder-decoder} model developed by Google AI, 
and GPT-4o-mini \citep{openai2024gpt4ocard}, a larger, state-of-the-art \emph{decoder-only} model developed by OpenAI.    
Importantly, our goal is not to teach general reasoning through fine-tuning, but to adapt models for specific tasks---premise selection and proof by contradiction---while improvements in reasoning skills may still arise.
We therefore view our results as reflecting the overall reasoning capabilities of pre-trained and fine-tuned models.
In this setting, for both tasks the input consists of a complete knowledge base paired with a hypothesis to be proven, while the output depends on the task: in premise selection, it is the subset of premises required to derive the hypothesis; 
in proof by contradiction, it is a formula enabling a type (iii) derivation. In both cases, the models process inputs and produce outputs as plain text sequences.    

\paragraph{Generalization Criteria}
We investigate two aspects of generalization in our framework: \emph{compositionality}, the capacity to deconstruct complex structures into simpler components, 
and \emph{recursiveness}, the ability to iteratively combine simpler structures to construct more complex ones. 
We operationalize these concepts by systematically excluding syllogisms with short and long \emph{A-chains} during training and evaluating model performance on them exclusively at test time. 
More precisely (see Figure~\ref{fig:com_rec_example}), a model is said to exhibit compositional generalization if it can correctly infer syllogisms with shorter \emph{A-chains} than 
those encountered during training. 
Conversely, it demonstrates recursive generalization if it successfully predicts syllogisms with longer \emph{A-chains} than those used in training.
To implement this evaluation, we excluded from the training set the five shortest and five longest \emph{A-chain} lengths for each syllogism type.

We first trained models without restrictions on \emph{A-chain} lengths and optimized their accuracy to establish overall performance. 
These \emph{overall} models serve as the baseline for evaluating generalization, as the \emph{compositional} and \emph{recursive} 
variants use the same training configuration, differing only in the exclusion of syllogisms with \emph{A-chain} in the designated length ranges.

\begin{figure}
    \centering    
    \tikzset{>=latex}    
    \scalebox{0.9}{\begin{tikzpicture}
        \node (x11) at (-0.5,0) {$a$};
        \node (x21) at (-0.5,1.5) {$b$};
        \node (x31) at (1,0) {$c$};
        \node (x41) at (1,1.5) {$d$};

        \node (x12) at (4.5,0) {$a$};
        \node (x22) at (4.5,1.5) {$b$};
        \node (x32) at (4.5,3) {$c$};
        \node (x42) at (6,0) {$d$};
        \node (x52) at (6,1.5) {$e$};
        \node (x62) at (6,3) {$f$};
        
        \draw[thick, Acolor, ->] (x11) -- (x21) node[midway,left] {$A$};
        \draw[thick, Acolor, ->] (x31) -- (x41) node[midway,right] {$A$};
        
        \draw[thick, Acolor, ->] (x12) -- (x22) node[midway,left] {$A$};
        \draw[thick, Acolor, ->] (x22) -- (x32) node[midway,left] {$A$};
        \draw[thick, Acolor, ->] (x42) -- (x52) node[midway,right] {$A$};
        \draw[thick, Acolor, ->] (x52) -- (x62) node[midway,right] {$A$};
                
        \draw[thick, Ocolor, ->] (x11) -- (x41);
        \draw[thick, dashed, Ocolor, ->] (x21) -- (x31); 
        \node (l1) at (0.25,1.2) {\color{Ocolor} $O$};						
        
        \draw[thick, Ocolor, ->] (x12) -- (x62);
        \draw[thick, dashed, Ocolor, ->] (x32) -- (x42); 
        \node (l2) at (5.25,2.2) {\color{Ocolor} $O$};

        \node (n1) at (0.2,-0.5) {$(1)$ short inference};
        \node (n2) at (5.2,-0.5) {$(2)$ long inference};
    \end{tikzpicture}}
    \\[0.3cm]    
    \scalebox{0.85}{\begin{tabular}{l @{\hspace{1.3cm}} l}
        $H_1 = Obc$ & $H_2 = Ocd$ \\
        $\NP_1 = \{ Aab, Acd, Oad \}$ & $\NP_2 = \{ Aab, Abc, Ade, Aef, Oaf \}$ \\
        $F_1 = Oad$ & $F_2 = Oaf$
    \end{tabular}}
    \caption{Example of inference-length generalization for a syllogism of type $\tau_1$. 
    Given a knowledge base and a hypothesis $H$ (shown with dashed lines), models predict a subset of premises $\NP$ and a refutation formula $F$ 
    for the premise selection and proof by contradiction tasks, respectively. 
    A recursive model trained on (1) generalizes to (2), while a compositional model trained on (2) generalizes to (1).}
    \label{fig:com_rec_example}
\end{figure}

\paragraph{Data Specification}
Our dataset comprises 30 distinct synthetic knowledge bases for fine-tuning and an additional 30 for evaluation. 
To further assess the models’ capacity for recursive generalization, we constructed an extra set of 60 KBs, 
specifically designed to include a greater proportion of inferences involving longer \emph{A-chains}, which are typically underrepresented.
On average, each knowledge base contains 40 premises and 1333 valid hypotheses.
A breakdown of the mean number of hypotheses per task, categorized by syllogism type, is provided in Table~\ref{tab:avg_kbs}.
Note that the proof by contradiction task excludes hypotheses that can be derived using only proof types (i) and (ii).

\begin{table}[t]	
	\centering
	\scalebox{0.85}{\begin{tabular}{ l c c c c c c c c }
			\toprule
			\textbf{Task} & $\T_1$ & $\T_2$ & $\T_3$ & $\T_4$ & $\T_5$ & $\T_6$ & $\T_7$ & \textbf{Total} \\
			\midrule
			PS & 68 & 152 & 245 & 513 & 42 & 110 & 203 & 1333 \\
			PBC & 62 & -- & 245 & 361 & 42 & -- & 202 & 911 \\
			\bottomrule
	\end{tabular}}
    \caption{Average number of hypotheses per KB by syllogism type for premise selection (PS) and proof by contradiction (PBC)}
	\label{tab:avg_kbs}
\end{table}

\begin{table}[t]	
	\centering
	\scalebox{0.85}{\begin{tabular}{ l c c c }
		\toprule
		& \textbf{Training} & \textbf{Test (OVE/COM)} & \textbf{Test (REC)} \\
		\midrule
		Base KBs     & 30 & 30 & 60 \\
		Substitutions  & 10 & 3  & 3  \\
		Permutations   & 3  & 1  & 1  \\
		Augmented KBs    & 900 & 90 & 180 \\
        \midrule
		Total Hypotheses & $\sim$1.1M & $\sim$107K & $\sim$30K \\
		\bottomrule
	\end{tabular}}
    \caption{Dataset sizes and augmentation settings for the training and evaluation splits; OVE, COM, and REC denote overall, compositional, and recursive models}
	\label{tab:data_specification}
\end{table}

To prevent memorization and enhance generalization, each KB was augmented by applying multiple pseudoword substitutions, 
and for each substitution, random permutations of the premises were generated.
Table~\ref{tab:data_specification} summarizes the number of base and augmented KBs, as well as the total number of valid hypotheses for each split (training and test).
Premise permutations are not included in the test dataset, as models trained with this augmentation generalize perfectly to unseen permutations.

Following KB augmentation, we train all models by subsampling from the total number of hypotheses, stratified by syllogism type and \emph{A-chain} length. 
T5 required 80\% of the data for optimal performance, while GPT achieved comparable results with only 25\%, and further increases provided only marginal gains.
It is worth noting that 25\% still represents a substantial proportion: our GPT models were trained on an average of 100 million 
tokens\footnotemark\footnotetext{Fine-tuning cost: $\approx\$3.1\text{K}$ (excluding evaluation and preliminary testing)}, 
even though OpenAI suggests that smaller datasets may suffice for fine-tuning. 
This highlights the necessity of large-scale data for robust performance in logical reasoning tasks.

\section{Empirical Analysis of Generalization in LLMs}
\label{sec:analysis_of_generalization}
As described in the previous section, we construct datasets $\mathcal{D}_{\text{train}}$ and $\mathcal{D}_{\text{test}}$ for training and evaluating overall models. 
To study generalization, we derive modified training sets by excluding instances with either the shortest or longest inference lengths from $\mathcal{D}_{\text{train}}$. 
For evaluation, we construct corresponding test subsets with short and long inference lengths, denoted as $\mathcal{D}_{\text{test}}^{s}$ and $\mathcal{D}_{\text{test}}^{l}$, respectively.
We design five experimental settings with different training and evaluation splits (see Table~\ref{tab:training_test_split}). 
Three settings correspond to overall models (OVE), evaluated in-distribution on $\mathcal{D}_{\text{test}}$, $\mathcal{D}_{\text{test}}^{s}$, and $\mathcal{D}_{\text{test}}^{l}$, 
namely OVE-all, OVE-short, and OVE-long, respectively. 
The remaining two settings correspond to compositional (COM) and recursive (REC) models, evaluated out-of-distribution on $\mathcal{D}_{\text{test}}^{s}$ and $\mathcal{D}_{\text{test}}^{l}$, respectively.

Table~\ref{tab:main_accuracy} presents the average accuracy across all evaluated knowledge bases for each experimental setting, using T5 and GPT architectures 
fine-tuned on premise selection and proof by contradiction tasks. 
Each experiment was run three times for T5 and twice for GPT, and we report the best run to provide a reference for subsequent evaluation of hybrid models.

\begin{table}[t]    
    \centering
    \scalebox{0.83}{\begin{tabular}{ l l l }
        \toprule
        \textbf{Experiment} & \textbf{Training} & \textbf{Test} \\
        \midrule
        OVE-all   & $\mathcal{D}_{\text{train}}$ & $\mathcal{D}_{\text{test}}$ \\
        OVE-short & $\mathcal{D}_{\text{train}}$ & $\mathcal{D}_{\text{test}}^{s}$ \\
        OVE-long  & $\mathcal{D}_{\text{train}}$ & $\mathcal{D}_{\text{test}}^{l}$ \\
        \midrule
        COM                & $\mathcal{D}_{\text{train}} \setminus \mathcal{D}_{\text{train}}^{s}$ & $\mathcal{D}_{\text{test}}^{s}$ \\
        REC                & $\mathcal{D}_{\text{train}} \setminus \mathcal{D}_{\text{train}}^{l}$  & $\mathcal{D}_{\text{test}}^{l}$ \\
        \bottomrule
    \end{tabular}}
    \caption{Training and evaluation splits for each experimental setting}
    \label{tab:training_test_split}
\end{table}

\begin{table}	
	\centering	
	\scalebox{0.83}{\begin{tabular}{ l c c c c }
			\toprule
			\multirow{2}{4em}{\textbf{Experiment}} & \multicolumn{2}{c}{\textbf{Premise Selection}} & \multicolumn{2}{c}{\textbf{Proof By Contradiction}} \\
			& \textbf{T5} & \textbf{GPT} & \textbf{T5} & \textbf{GPT} \\
			\midrule
            OVE-all    & \msd{0.94}{0.05} & \msd{0.94}{0.05} & \msd{0.93}{0.04} & \msd{0.95}{0.04} \\
            OVE-short  & \msd{0.99}{0.01} & \msd{0.98}{0.01} & \msd{0.99}{0.01} & \msd{0.98}{0.02} \\
            OVE-long   & \msd{0.79}{0.17} & \msd{0.83}{0.15} & \msd{0.76}{0.15} & \msd{0.88}{0.15} \\
            \midrule
            COM                 & \msd{0.84}{0.04} & \msd{0.76}{0.04} & \msd{0.67}{0.05} & \msd{0.85}{0.05} \\
            REC                 & \msd{0.80}{0.15} & \msd{0.82}{0.15} & \msd{0.71}{0.18} & \msd{0.86}{0.17} \\
			\bottomrule
	\end{tabular}}		
    \caption{Accuracy scores for premise selection and proof by contradiction tasks (all experiments)}
	\label{tab:main_accuracy}
\end{table}

The OVE family of experiments is designed to characterize generalization to unseen KB structures under seen inference lengths. Results show that OVE-all achieves consistently high accuracy across all tasks and models, while OVE-short and OVE-long reveal a dependence on inference length. We then turn to the COM and REC settings, which test models on both unseen KB structures and unseen inference lengths. In the COM setting, performance drops substantially relative to OVE-short, where accuracy is near perfect. In contrast, in the REC setting, both T5 and GPT perform comparably to OVE-long, but remain below the performance observed in OVE-short.

\begin{figure}
	\centering
	\begin{tabular}{ c c }
		\multicolumn{2}{c}{\small Task: Premise selection} \\
		{\small Compositionality} & {\small Recursiveness} \\
		\includegraphics[width=0.47\linewidth]{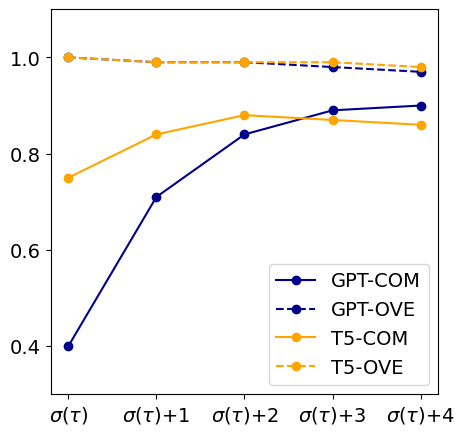} & \includegraphics[width=0.455\linewidth]{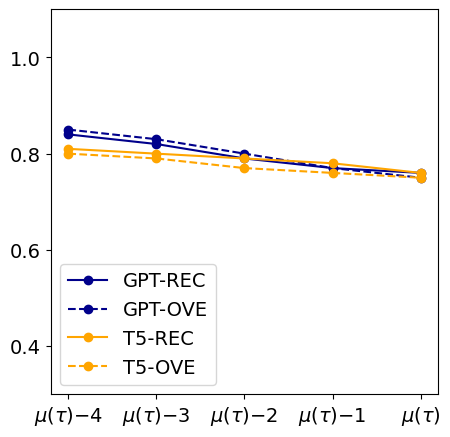} \hspace{5mm} \\
		\multicolumn{2}{c}{\small Task: Proof By Contradiction} \\
		{\small Compositionality} & {\small Recursiveness} \\
		\includegraphics[width=0.47\linewidth]{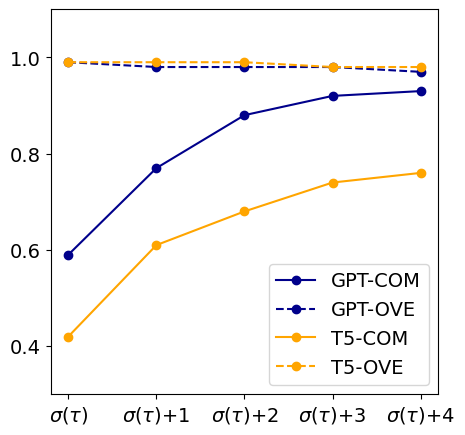} & \includegraphics[width=0.455\linewidth]{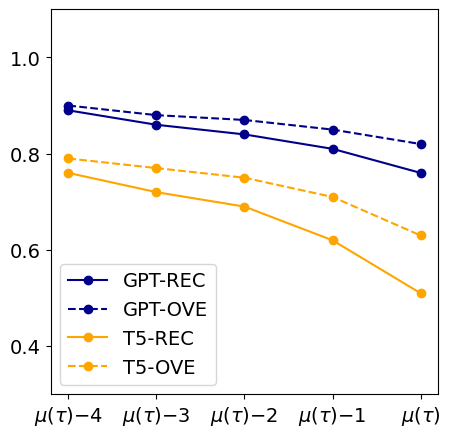} \hspace{5mm}
	\end{tabular}			
    \caption{Generalization performance of GPT and T5 architectures across the five shortest and five longest unseen \emph{A-chain} lengths. 
    The y-axis shows accuracy, and the x-axis shows unseen lengths. For each syllogism type $\T$, $\sigma(\T)$ and $\mu(\T)$ denote the shortest and longest lengths considered, respectively.}
	\label{fig:unseen_lengths}
\end{figure}

A more comprehensive analysis is provided in Figure \ref{fig:unseen_lengths}, which illustrates the accuracy of evaluated inferences across the five shortest and five longest unseen \emph{A-chain} lengths.
For a given syllogism of type $\T$, we denote the shortest evaluated length as $\sigma(\T)$ and the longest as $\mu(\T)$.
Solid lines in the plot represent compositional and recursive models, while dashed lines depict overall models. Note that overlapping lines would imply a perfect generalization. 
The latter is visible in the recursive evaluation for the premise selection task (top-right plot). However, a slight drop in the accuracy occurs as the lengths increase. 
The drop is more evident for the proof by contradiction task (bottom-right plot). This suggests that LLMs experience difficulties in generating long sequences of $A$-formulas, 
regardless of their presence during training.
Compositional experiments, on the other hand, exhibit a poor generalization and a steeper curve towards the shortest length, 
in contrast to overall models that can generate inferences involving shorter \emph{A-chains} almost perfectly.

\subsection{Analysis by Syllogism Type}
\label{sec:analysis_by_type}
We further analyze these experiments by considering each specific syllogism type (see Table~\ref{tab:syllogisms}) 
and decomposing the results in Table~\ref{tab:main_accuracy} into a detailed breakdown of accuracy scores for each reasoning task: 
premise selection and proof by contradiction.

\paragraph{Premise Selection Task}
The outcomes of the premise selection task are reported in Tables \ref{tab:t5_ps_types_accuracy} and \ref{tab:gpt_ps_types_accuracy} for the T5 and GPT models, respectively. 
Notably, type $\T_2$---the structurally simplest inference embedded within all other types---consistently achieved the highest accuracy across all experimental settings, 
demonstrating near-perfect generalization, with the unique exception of GPT in the compositionality experiment. 	
In contrast, types $\T_1$ and $\T_5$ achieved the lowest accuracies for T5 across the generalization tasks, whereas for GPT, poor performance was observed only on type $\T_1$.

\begin{table}	
	\centering
	\scalebox{0.78}{\begin{tabular}{lccccc}
		\toprule
		\textbf{$\T$} & \textbf{OVE-all} & \textbf{OVE-short} & \textbf{OVE-long} & \textbf{COM} & \textbf{REC} \\
		\midrule
		1 & \msd{0.90}{0.10} & \msd{0.96}{0.05} & \msd{0.58}{0.38} & \msd{0.35}{0.11} & \msd{0.59}{0.35} \\
		2 & \msd{1.00}{0.01} & \msd{1.00}{0.00} & \msd{0.97}{0.09} & \msd{0.99}{0.01} & \msd{0.96}{0.08} \\
		3 & \msd{0.85}{0.13} & \msd{0.96}{0.05} & \msd{0.61}{0.37} & \msd{0.88}{0.06} & \msd{0.66}{0.33} \\
		4 & \msd{0.97}{0.03} & \msd{1.00}{0.01} & \msd{0.77}{0.21} & \msd{0.87}{0.06} & \msd{0.74}{0.23} \\
		5 & \msd{0.72}{0.28} & \msd{0.71}{0.36} & \msd{0.65}{0.41} & \msd{0.49}{0.37} & \msd{0.69}{0.38} \\
		6 & \msd{0.97}{0.08} & \msd{0.99}{0.03} & \msd{0.84}{0.33} & \msd{0.90}{0.05} & \msd{0.88}{0.28} \\
		7 & \msd{0.96}{0.08} & \msd{1.00}{0.01} & \msd{0.76}{0.31} & \msd{0.77}{0.12} & \msd{0.80}{0.28} \\
		\bottomrule
	\end{tabular}}
    \caption{T5 accuracy scores by type (premise selection)}
	\label{tab:t5_ps_types_accuracy}
\end{table}

\begin{table}	
	\centering
	\scalebox{0.78}{\begin{tabular}{lccccc}
		\toprule
		\textbf{$\T$} & \textbf{OVE-all} & \textbf{OVE-short} & \textbf{OVE-long} & \textbf{COM} & \textbf{REC} \\
		\midrule
		1 & \msd{0.84}{0.15} & \msd{0.91}{0.10} & \msd{0.54}{0.36} & \msd{0.20}{0.11} & \msd{0.52}{0.38} \\
		2 & \msd{1.00}{0.01} & \msd{1.00}{0.00} & \msd{0.98}{0.06} & \msd{0.89}{0.04} & \msd{0.97}{0.06} \\
		3 & \msd{0.87}{0.13} & \msd{0.98}{0.04} & \msd{0.67}{0.36} & \msd{0.87}{0.07} & \msd{0.68}{0.35} \\
		4 & \msd{0.97}{0.04} & \msd{0.99}{0.02} & \msd{0.81}{0.22} & \msd{0.76}{0.05} & \msd{0.78}{0.24} \\
		5 & \msd{0.84}{0.23} & \msd{0.93}{0.17} & \msd{0.70}{0.38} & \msd{0.93}{0.16} & \msd{0.68}{0.39} \\
		6 & \msd{0.99}{0.04} & \msd{1.00}{0.02} & \msd{0.94}{0.19} & \msd{0.92}{0.02} & \msd{0.92}{0.23} \\
		7 & \msd{0.96}{0.09} & \msd{0.99}{0.03} & \msd{0.86}{0.25} & \msd{0.71}{0.12} & \msd{0.86}{0.24} \\
		\bottomrule
	\end{tabular}}
    \caption{GPT accuracy scores by type (premise selection)}
	\label{tab:gpt_ps_types_accuracy}
\end{table}

\paragraph{Proof By Contradiction Task}
Tables \ref{tab:t5_pbc_types_accuracy} and \ref{tab:gpt_pbc_types_accuracy} present the results for proof by contradiction.
For this reasoning task---where types $\T_2$ and $\T_6$ were not applicable---type $\T_7$ emerged as the best-performing inference type
on both architectures, achieving a near-perfect generalization in the case of GPT. 
Overall, this task appeared to be easier for GPT than for T5, which, with few exceptions, struggled to generalize effectively. 
GPT, by contrast, demonstrated consistently strong performance across all inference types, including the particularly challenging type $\T_1$. 

\begin{table}	
	\centering
	\scalebox{0.78}{\begin{tabular}{lccccc}
		\toprule
		\textbf{$\T$} & \textbf{OVE-all} & \textbf{OVE-short} & \textbf{OVE-long} & \textbf{COM} & \textbf{REC} \\
		\midrule
		1 & \msd{0.97}{0.05} & \msd{0.98}{0.04} & \msd{0.88}{0.21} & \msd{0.23}{0.12} & \msd{0.82}{0.27} \\
		3 & \msd{0.88}{0.09} & \msd{0.98}{0.04} & \msd{0.54}{0.33} & \msd{0.95}{0.05} & \msd{0.37}{0.33} \\
		4 & \msd{0.97}{0.03} & \msd{1.00}{0.00} & \msd{0.78}{0.20} & \msd{0.66}{0.06} & \msd{0.77}{0.22} \\
		5 & \msd{0.78}{0.29} & \msd{0.68}{0.39} & \msd{0.78}{0.33} & \msd{0.36}{0.36} & \msd{0.84}{0.31} \\
		7 & \msd{0.95}{0.07} & \msd{0.99}{0.03} & \msd{0.85}{0.25} & \msd{0.83}{0.12} & \msd{0.80}{0.29} \\
		\bottomrule
	\end{tabular}}
    \caption{T5 accuracy scores by type (proof by contradiction)}
	\label{tab:t5_pbc_types_accuracy}
\end{table}

\begin{table}	
	\centering
	\scalebox{0.78}{\begin{tabular}{lccccc}
		\toprule
		\textbf{$\T$} & \textbf{OVE-all} & \textbf{OVE-short} & \textbf{OVE-long} & \textbf{COM} & \textbf{REC} \\
		\midrule
		1 & \msd{0.97}{0.06} & \msd{0.96}{0.07} & \msd{0.93}{0.20} & \msd{0.90}{0.14} & \msd{0.90}{0.24} \\
		3 & \msd{0.92}{0.12} & \msd{0.97}{0.08} & \msd{0.81}{0.30} & \msd{0.97}{0.05} & \msd{0.74}{0.34} \\
		4 & \msd{0.97}{0.04} & \msd{0.99}{0.02} & \msd{0.87}{0.19} & \msd{0.78}{0.05} & \msd{0.81}{0.23} \\
		5 & \msd{0.85}{0.30} & \msd{0.78}{0.37} & \msd{0.96}{0.16} & \msd{0.73}{0.37} & \msd{0.93}{0.22} \\
		7 & \msd{0.97}{0.07} & \msd{0.98}{0.06} & \msd{0.92}{0.19} & \msd{0.98}{0.04} & \msd{0.95}{0.15} \\
		\bottomrule
	\end{tabular}}
    \caption{GPT accuracy scores by type (proof by contradiction)}
	\label{tab:gpt_pbc_types_accuracy}
\end{table}

\subsection{Analysis of Incorrect Predictions}
\label{sec:analysis_incorrect_predictions}
We first analyze the types of errors that arise in the premise selection task. In particular, we distinguish between cases where the model includes unnecessary premises, still allowing the construction of valid proofs, 
and cases where there are missing premises, representing true prediction failures.

\paragraph{Unnecessary Premises}
We have found that models occasionally predict unnecessary premises---i.e., premises that are not strictly required to derive the target hypothesis. 
While such predictions are technically incorrect with respect to minimal inferences, they nonetheless allow to construct a valid proof of the hypothesis. 
Therefore, they remain acceptable inputs for the symbolic prover.

We performed an analysis by treating such predictions as correct and report the resulting accuracy gain, 
defined as the difference between non-minimal and minimal accuracies in the generalization experiments. 
Tables \ref{tab:gain_nm_com} and \ref{tab:gain_nm_rec} present the results for the compositional and recursive settings, respectively. 
We additionally report the mean and standard deviation of the number of unnecessary premises predicted by each model. 
In the compositional setting, T5 shows only a marginal improvement of $\approx 2\%$, whereas GPT exhibits a substantial gain of $\approx 9\%$. 
In the recursive setting, T5 shows no improvement, while GPT achieves a modest gain of $\approx 2\%$.
Regarding unnecessary premises, in the compositional setting both T5 and GPT predict a similar average number ($\approx 5$), 
whereas in the recursive setting T5 predicts slightly more ($\approx 6$) than GPT ($\approx 4$).

\begin{table}    
    \centering
    \scalebox{0.85}{
        \begin{tabular}{l c c c c}
        \toprule
        \multirow{2}{4em}{\textbf{Length}} & \multicolumn{2}{c}{\textbf{T5}} & \multicolumn{2}{c}{\textbf{GPT}} \\
        & \textbf{Gain} & \textbf{\# Unnec. Prem.} & \textbf{Gain} & \textbf{\# Unnec. Prem.} \\
        \midrule
        $\sigma(\T)$   & 0.08 & \msd{5.55}{4.66} & 0.26 & \msd{5.51}{3.53} \\
        $\sigma(\T)+1$ & 0.02 & \msd{5.77}{4.11} & 0.10 & \msd{5.33}{3.86} \\
        $\sigma(\T)+2$ & 0.00 & \msd{4.76}{4.10} & 0.04 & \msd{5.28}{3.63} \\
        $\sigma(\T)+3$ & 0.01 & \msd{4.40}{3.79} & 0.02 & \msd{5.31}{3.78} \\
        $\sigma(\T)+4$ & 0.00 & \msd{5.22}{4.87} & 0.02 & \msd{5.30}{3.94} \\
        \midrule
        Total          & 0.02 & \msd{5.44}{4.48} & 0.09 & \msd{5.42}{3.66} \\
        \bottomrule
    \end{tabular}}
    \caption{Accuracy gain from non-minimal proofs and number of unnecessary premises for shorter unseen lengths (COM)}
    \label{tab:gain_nm_com}
\end{table}

\begin{table}    
    \centering
    \scalebox{0.85}{
        \begin{tabular}{l c c c c}
        \toprule
        \multirow{2}{4em}{\textbf{Length}} & \multicolumn{2}{c}{\textbf{T5}} & \multicolumn{2}{c}{\textbf{GPT}} \\
        & \textbf{Gain} & \textbf{\# Unnec. Prem.} & \textbf{Gain} & \textbf{\# Unnec. Prem.} \\
        \midrule
        $\mu(\T)-4$ & 0.00 & \msd{5.21}{3.48} & 0.02 & \msd{4.22}{4.22} \\
        $\mu(\T)-3$ & 0.01 & \msd{6.86}{3.58} & 0.02 & \msd{4.21}{4.16} \\
        $\mu(\T)-2$ & 0.00 & \msd{7.46}{3.77} & 0.02 & \msd{3.48}{3.47} \\
        $\mu(\T)-1$ & 0.00 & \msd{2.75}{1.30} & 0.02 & \msd{3.46}{3.69} \\
        $\mu(\T)$   & 0.00 & \msd{5.67}{1.25} & 0.01 & \msd{3.86}{4.69} \\
        \midrule
        Total       & 0.00 & \msd{6.06}{3.63} & 0.02 & \msd{3.98}{4.04} \\
        \bottomrule
    \end{tabular}}
    \caption{Accuracy gain from non-minimal proofs and number of unnecessary premises for longer unseen lengths (REC)}
    \label{tab:gain_nm_rec}
\end{table}

\paragraph{Missing Premises}	
Finally, we examined incorrect model predictions that did not involve the inclusion of unnecessary premises, 
in order to better understand the inferential patterns being captured. In all such cases, the models consistently generated well-formed formulas. 
Therefore, the observed errors can be attributed to the incorrect selection or construction of syllogistic formulas, rather than to syntactic malformation.

To evaluate semantic errors in the premise selection task, we considered three key aspects: 
\emph{term overlap}, assessing whether the terms appearing in the hypothesis were also present in the predicted premises, which would indicate a basic understanding of syllogistic rules; 
\emph{premise validity}, examining whether the predicted premises were sourced from the knowledge base; and 
\emph{term validity}, verifying whether the terms within the predicted premises were restricted to those found in the knowledge base. 
These latter checks ensure that the models are not generating fabricated content.

\begin{table}	
	\centering	
	\scalebox{0.85}{\begin{tabular}{ l l r r r }
			\toprule
			\textbf{Model} & \textbf{Exp.} & \textbf{Term overlap} & \textbf{Prem. valid.} & \textbf{Term val.} \\
			\midrule
			\multirow{3}{*}{T5} & OVE & 0.77 & 0.35 & 0.92 \\
			& COM & 0.46 & 0.25 & 0.90 \\
			& REC & 0.63 & 0.18 & 0.89 \\
			\midrule
			\multirow{3}{*}{GPT} & OVE & 0.94 & 0.20 & 0.92 \\
			& COM & 0.71 & 0.31 & 0.94 \\
			& REC & 0.92 & 0.32 & 0.92 \\
			\bottomrule
	\end{tabular}}
    \caption{Semantic validity for the premise selection task}
	\label{tab:err_ps}
\end{table}

Table \ref{tab:err_ps} presents a summary of semantic validity by reporting the proportion of cases, across all experiments, in which each criterion was satisfied for both T5 and GPT models. 
These proportions were calculated relative to the set of incorrect predictions. Interestingly, the results indicate that the models generate fabricated premises. 
A closer analysis suggests that this issue may stem from confusion between syllogisms whose hypotheses share the same formula type---specifically, 
among $O$-formulas in inference types $\T_1$, $\T_3$, and $\T_5$, and among $I$-formulas in types $\T_4$ and $\T_7$.
That is, the models appear to erroneously construct an incorrect type of syllogism, thereby generating fabricated premises in an attempt to force a valid inference.
We illustrate this phenomenon with two representative cases:

\begin{itemize}
    \item First, consider syllogisms of type $\T_1$. To derive $Obc$, the required premises are $\{ \Achain{a}{b}, \Achain{c}{d}, Oad \}$.
            However, the model attempts to construct a proof of type $\T_3$ instead,  which requires the premises $\{ \Achain{a}{b}, \Achain{c}{d}, \Achain{a}{e}, Ede \}$.
            In doing so, the model predicts $Ede$ instead of the required $Oad$ and assumes the existence of $\Achain{a}{e}$.             
    \item Second, consider syllogisms of type $\T_4$, where the goal is to derive $Ibc$ from the premises $\{ \Achain{a}{b}, \Achain{a}{c} \}$.
            Here, the model sometimes attempts to apply type $\T_7$, which has the form $\{ \Achain{a}{b}, \Achain{c}{d}, Iac \} \vdash Ibd$.
            In this case, the model predicts a formula of the form $Iax$ and assumes a formula $\Achain{x}{c}{}$ that does not exist.
\end{itemize}

As a result, in both cases, the model may fabricate premises in order to complete the incorrectly assumed proof structure.

\paragraph{Incorrect Formulas in the Proof by Contradiction Task}
Similarly, in the proof by contradiction task, all incorrect predictions were syntactically well-formed. 
While the criteria concerning term overlap with the hypothesis and the validity of premises are not directly applicable in this setting, 
it is noteworthy that, in all instances, the terms used in the incorrectly predicted formulas were contained within the vocabulary of the knowledge base (\emph{term validity}).

\section{Neuro-Symbolic Reasoning Framework}
\label{sec:hybrid_models}
To formalize the integration of symbolic and connectionist components, we represent the neural networks as functions that map a hypothesis and a knowledge base to a set of syllogistic formulas:
$$
\PS : (H, \KB) \mapsto 2^{\mathcal{F}} \quad\quad \PBC : (H, \KB) \mapsto 2^{\mathcal{F}}.
$$
Here, $\PS$ denotes the \emph{premise selection} model, while $\PBC$ denotes the \emph{proof by contradiction} model. Both neural components are designed to provide assistance to a symbolic 
prover. An overview of the resulting hybrid framework is illustrated in Figure \ref{fig:hybrid_architecture}. 
In the remainder of this section, we present a simple algorithm that implements this integration, followed by an experimental evaluation assessing the behavior and performance of 
the proposed hybrid models.

\subsection{Integration of Symbolic and Connectionist Components}
The hybrid model reuses the symbolic algorithms from Section~\ref{sec:symbolic_component} while incorporating predictions from neural assistants 
to restrict the search space and reduce non-deterministic exploration during inference.
Algorithm~\ref{alg1:hybrid_syllogistic_prover} summarizes the resulting hybrid syllogistic prover. Given a knowledge base $\KB$ and a hypothesis $H$, 
the neural components $\PS$ and $\PBC$ first predict the necessary premises to prove $H$, and formulas to prove $H$ by contradiction, respectively. 
These predictions are used to initialize the symbolic search (lines 1-2). In particular, the set of contradictory pairs $\Cont$ (cf. Algorithm~\ref{alg3:pbc}, line 1) 
is initialized with $\textsc{cp}(\Fpbc) = \{ (F, \overline{F}) \,|\, F \in \Fpbc \}$ so that candidate pairs suggested by the neural assistant are explored first (line 3).
The symbolic prover (Algorithm~\ref{alg1:syllogistic_prover}) is then executed on a restricted subset $\NP \subseteq \KB$ (line 4). 
If no proof is found, the prover is called again using the full knowledge base (line 8), thereby reverting to the purely symbolic setting.

Two properties of the framework are worth noting. First, if the \emph{premise selection} model $\PS$ fails to provide the premises necessary for deriving the hypothesis, 
the procedure restarts without restrictions, ensuring completeness at the cost of additional computation. 
Second, the \emph{proof by contradiction} model $\PBC$ may affect only the order in which contradictory pairs are explored; even if its predictions are ineffective, 
the non-deterministic nature of the search guarantees no impact on correctness and does not require additional computation. 
Thus, neural assistance influences only efficiency, not correctness, and the hybrid prover is guaranteed to terminate and to remain sound and complete regardless of neural performance.

\begin{algorithm}
	\caption{Hybrid Syllogistic Prover}
	\label{alg1:hybrid_syllogistic_prover}
    \begin{algorithmic}[1]
        \Require A hypothesis $H$ and a knowledge base $\KB$.
        \Ensure A proof $\D$ of $H$ from $\KB$ (if $H$ is valid).        
        
        \State $\NP \gets \PS(H, \KB)$ 
        \State $\Fpbc \gets \PBC(H, \KB)$ 
        \State $\Cont \gets \langle \Call{cp}{\Fpbc} \rangle \Vert \langle$ \Call{all\_pairs}{$\Q, \X$} $\setminus \Call{cp}{\Fpbc} \rangle$ 
        
        \State $\D \gets$ \Call{Syllogistic Prover}{$H, \NP$} 
        
        \If{$\D \neq \texttt{False}$}
            \State \textbf{return} $\D$
        \EndIf
    
        \State \textbf{return} \Call{Syllogistic Prover}{$H, \KB$} 
    \end{algorithmic}
\end{algorithm}

\subsection{Experimental Evaluation}
We investigated three variants of hybrid models, each incorporating neural components trained under different generalization regimes---namely, overall, compositional, and recursive---and 
evaluated their performance relative to a purely symbolic baseline to assess the impact of the neural assistants.

\paragraph{Sampling Procedure} We randomly selected just over 2000 samples from 30 distinct knowledge bases within the test dataset.
From each knowledge base, we selected 10 syllogisms of each type, comprising 5 instances featuring longer \emph{A-chains} and 5 with shorter \emph{A-chains}. 
To ensure a sufficient level of inference complexity, we excluded trivial proofs by retaining only those syllogisms in which the \emph{A-chains} length is greater than or equal to 2.

\begin{figure}[t]
	\centering
	\includegraphics[width=1\linewidth]{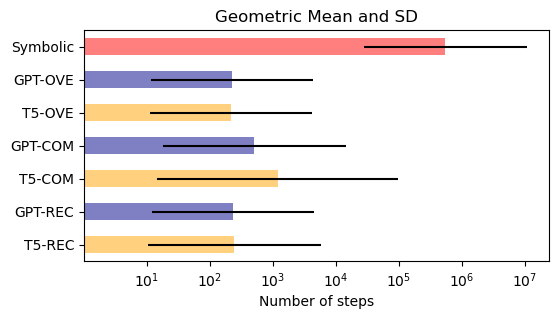}
    \caption{Geometric mean and standard deviation of the number of steps for the symbolic and hybrid models using different assistants trained on GPT and T5. OVE, COM, and REC denote overall, compositional, and recursive models, respectively.}
	\label{fig:t5_hm_mean_sd}
\end{figure}

\paragraph{Symbolic vs. Hybrid Evaluation}
To compare the efficiency of symbolic and hybrid models, we measured the number of steps required to complete the proof of a valid hypothesis, 
where each step corresponds to a single invocation of the recursive function \textsc{drv} by the prover. 
Due to the non-deterministic nature of the symbolic component, each experiment is executed five times for each model configuration.
Moreover, we use logarithmic notation and the geometric mean---due to the potentially wide variability--- to represent step counts (see Figure \ref{fig:t5_hm_mean_sd}).

On average, the symbolic model requires approximately $10^{5.7}$ steps to complete a proof. In contrast, hybrid models require substantially fewer steps.
In particular, models incorporating overall and recursive neural assistants, exhibit comparable performance, and they need only around $10^{2.4}$ steps across both LLM architectures. 
This corresponds to a reduction of approximately three orders of magnitude. Hybrid configurations assisted by compositional models require slightly more steps, approximately
$10^{2.7}$ for the GPT-based model and $10^{3.1}$ for the T5-based model. This outcome is not unexpected, as they exhibit a drop in accuracy relative to their overall and recursive counterparts. 
However, their use in hybrid configurations does not result in a significant increase in derivation steps, indicating robustness when assisting the prover.

\section{Conclusions}	
\label{sec:conclusions}
Our main findings focus on the performance of Large Language Models (LLMs) and their role in assisting automated provers. From a semantic perspective, LLMs struggle to fully grasp logical reasoning. Our generalization experiments highlight a significant gap between recursiveness and compositionality, indicating a need for a deeper theoretical understanding of this issue.
One possible explanation is that recursiveness relies on repetitive pattern matching, whereas compositionality requires structural decomposition---a process that transformer attention mechanisms may not naturally support.
To better understand this gap, we provide a more fine-grained analysis across different syllogism types
where we observed notable differences in performance and generalization. 
In particular, language models can reason about single \emph{A-chains} (syllogisms of type $\T_2$), 
whereas types combining multiple \emph{A-chains} with additional syllogistic formulas pose greater challenges.
These findings open a research agenda for understanding how different reasoning building blocks affect model performance, including in more complex fragments of logic, where new generalization patterns may appear.

We performed a systematic analysis to investigate the models’ incorrect outputs.
Two patterns of interest were noted: models occasionally predict unnecessary premises while still including the correct ones, 
and sometimes confuse syllogisms whose hypotheses share the same formula type, producing fabricated premises. 
These observations motivate further investigation into the sources of such errors and the development of methods to mitigate them.

We consider a relatively small encoder–decoder model (T5), chosen for its efficiency and strong overall performance on our tasks, alongside a substantially larger decoder-only model (GPT), selected to evaluate a state-of-the-art LLM.
GPT shows greater efficiency, converging with less data, though both models achieve comparable performance. This suggests that scaling to larger models alone may not be sufficient to overcome the challenges posed by these reasoning tasks.
Nevertheless, our results demonstrate that neither the limitations in generalization nor model size prevent LLMs from effectively assisting symbolic provers.
On the contrary, this assistance fosters a collaborative relationship between connectionist and symbolic models. While connectionist models can simplify and expedite tasks, symbolic models can still complete them when necessary, making hybrid models an important area for further investigation. 

This study focuses on syllogistic logic, one of the simplest fragments of formal reasoning. Despite its limited expressive power, it already exposes non-trivial challenges for generalization. 
Future work will aim to explain why certain reasoning types are learnable while others are not, and to clarify whether these limitations are grounded in structural aspects of logical composition. 
To further investigate this question, we will extend this analysis to richer fragments, e.g., \cite{Pratt-Hartmann2004-PRAFOL} or suitable fragments of modal logic. 
Moreover, our modular approach could provide practical solutions for developing computationally efficient provers for these logics, 
forming part of a broader effort to determine where the boundary of tractability lies for neural and neuro-symbolic reasoners.

\section*{Acknowledgments}
This research was funded by the National Science Center, Poland under the OPUS call [grant 2020/37/B/HS1/04220]. 
We gratefully acknowledge Polish high-performance computing infrastructure PLGrid (HPC Center: ACK Cyfronet AGH) for providing computer facilities and support within computational grant no. PLG/2024/017165.

\section*{AI Declaration}
The authors have not employed any Generative AI tools.

\bibliographystyle{kr}
\bibliography{references}

\end{document}